# Pericoronary adipose tissue (PCAT) feature analysis in CT calcium score images with comparison to coronary CTA

Yingnan Song,[a] BS; Hao Wu,[a] PhD; Juhwan Lee,[a] PhD; Justin Kim,[a] BS; Ammar Hoori,[a] PhD; Tao Hu,[a] MS; Vladislav Zimin,[b] MD; Mohamed Makhlouf,[b] MD; Sadeer Al-Kindi,[b] MD; Sanjay Rajagopalan,[b] MD; Chun-Ho Yun,[c] MD; Chung-Lieh Hung,[d] MD; David L. Wilson,[a,e] PhD

[a] Department of Biomedical Engineering, Case Western Reserve University, Cleveland, OH, USA
[b] Harrington Heart and Vascular Institute, University Hospitals Cleveland Medical Center, Cleveland, OH, USA
[c] Division of Radiology, Department of Internal Medicine, MacKay Memorial Hospital, Taipei, Taiwan
[d] Division of Cardiology, Department of Internal Medicine, MacKay Memorial Hospital, Taipei, Taiwan
[e] Department of Radiology, Case Western Reserve University, Cleveland, OH, USA

**Abstract:**

Objective: We investigated the feasibility and advantages of using non-contrast CT calcium score (CTCS) images to assess pericoronary adipose tissue (PCAT) and its association with major adverse cardiovascular events (MACE).

Background: PCAT features from coronary CTA (CCTA) have been shown to be associated with cardiovascular risk but are potentially confounded by iodine. If PCAT in CTCS images can be similarly analyzed, it would avoid this issue and enable its inclusion in formal risk assessment from readily available, low-cost CTCS images.

Methods: To identify coronaries in CTCS images that have subtle visual evidence of vessels, we registered CTCS with paired CCTA images having coronary labels. We developed a novel "axial-disk" method giving regions for analyzing PCAT features in three main coronary arteries. We analyzed novel hand-crafted and radiomic features using univariate and multivariate logistic regression prediction of MACE and compared results against those from CCTA.

Results: Registration accuracy was sufficient to enable the identification of PCAT regions in CTCS images. Motion or beam hardening artifacts were often present in "high-contrast" CCTA but not CTCS. Mean HU and volume were increased in both CTCS and CCTA for MACE group. There were significant positive correlations between some CTCS and CCTA features, suggesting that similar characteristics were obtained. Using hand-crafted/radiomics from CTCS and CCTA, AUCs were 0.82/0.79 and 0.83/0.77 respectively, while Agatston gave AUC=0.73.

Conclusions: Preliminarily, PCAT features can be assessed from three main coronary arteries in non-contrast CTCS images with performance characteristics that are at the very least comparable to CCTA.

**Abbreviation list:**

MACE = major adverse cardiovascular event
CCTP = cardiac computed tomography perfusion
CTCS = computed tomography calcium score

CCTA = coronary computed tomography angiography
PCAT = pericoronary adipose tissue
CAC = coronary artery calcium
MPR = multiplanar reformatting

**Introduction:**

Over an extensive period, numerous models have been constructed to forecast cardiovascular risk, and these have been employed to provide patients with informed guidance and tailor interventions accordingly. Detection and quantification of coronary artery calcium (CAC) in computed tomography (CT) has been used as a powerful risk marker to predict major adverse cardiovascular events (1–3). Despite its widespread use, there are continuing limitations including attenuation of predictive accuracy in certain populations and at the extremes of CAC (4), suggesting room for improvement.

Recently, there has been a notable focus on pericoronary adipose tissue (PCAT) as assessed in coronary computed tomography angiography (CCTA) images as a risk factor for major adverse cardiovascular events (MACE). A variety of pathways have been suggested where PCAT inflammation may be involved in local stimulation of atherosclerotic plaque formation (5,6). Beyond volumes, characteristics of PCAT in CT are associated with an inflammatory signature and cardiovascular risk. The "outside-in" theory implicates inflamed adipocytes in PCAT in the production of adipocytokines which through its effects on adventitia lead to atherosclerosis (7). Numerous studies have shown the importance of PCAT in predicting inflammatory segments of the vessel wall (8,9). Fat depots with high lipid content have lower attenuation in CT, while more aqueous adipose tissues has a higher HU value than adipose tissue (10). Fat attenuation index (FAI), a metric related to CCTA mean HU values in a 3D distribution of tissue around the coronary artery region have been linked to inflammation, the association has proven from biopsy samples taken from patients undergoing cardiac surgery (11). CCTA-based radiomic profiling of coronary artery PCAT detects perivascular structural remodeling associated with coronary artery disease and improves cardiovascular risk prediction (12,13).

Regarding CCTA analysis of PCAT, the variable presence of iodine could be a confound, especially regarding average HU values and the size of PCAT volume. Recent studies have shown that the impact of iodine contrast in CCTA PCAT assessment, demonstrated as increased HU value, leading to false positive of fat inflammation detection. Almeida et al. found lower mean HU value and increased volume before contrast injection when using the traditional fat window (-190 HU, -30 HU) (14). Our group investigated PCAT in cardiac CT perfusion (CCTP) studies, demonstrated dynamic PCAT enhancement, and assessed perfusion blood flow in PCAT (15). The presence of iodine can confound HU attenuation, volume, and radiomic features, all of which can further depend on the timing of the CCTA acquisition and presence of obstructive disease. In addition, coronary artery motion, scan parameters (e.g., kVp), beam hardening effects from contrast agent, and location and depth from the body surface, may all impede accurate PCAT assessment in CCTA (16).

An alternative to PCAT analysis in CCTA images is to use non-contrast CT calcium score (CTCS) images. The use of CTCS images presents an advantage as there are exceptionally large cohorts of inexpensive CTCS images, enabling big data analysis for machine learning. Current studies have extracted PCAT from CTCS images based on manual segmentation. Jiang et al. manually segmented RCA PCAT and demonstrated that radiomics extracted are informative to identify non-calcified, vulnerable plaques (17). Takahashi et al. placed a 15 x 15 $mm^2$ region of interest to extract RCA PCAT, mean HU was significantly associated high risk plaque (18). Notably, these studies using manual segmentation on the RCA only, which tends to have better visibility than the LAD and LCX in the non-contrast images. It will be important to extend PCAT analysis to other arteries.

In this study, we investigated the feasibility and advantages of using non-contrast CTCS images to assess features of PCAT and determine their association to MACE. We developed a specialized processing pipeline to analyze PCAT associated with LAD and LCX, and, in addition, RCA, which has not been previously analyzed. Using a non-rigid method, we registered CCTA images to paired CTCS images and transferred coronary artery labels from CCTA to the CTCS images for PCAT extraction. As CTCS images have relative thick slice thickness comparing to CCTA ones, we created an axial-disk method for analysis of PCAT hand-crafted and radiomics library features. We compared feature results from pairs of CTCS and iodine-containing CCTA images. Finally, we performed preliminary analysis of MACE prediction comparing CTCS to CCTA and Agatston score.

**Methods**

**Data acquisition**

This study was approved as a retrospective study of de-identified images by our local institutional review board. Images were acquired starting in 2013 at Mackay memorial hospital, Taiwan, and shared under a data use agreement. The population consisted of 83 consecutive patients with suspected coronary artery disease (CAD) who underwent both CTCS (120-kVp, 30-mAs, 3-mm slice thickness)) from a dual source scanner (Siemens SOMATOM and CCTA (100-kVp, 600-mAs, 0.75-mm slice thickness Definition Flash). CCTA and CTCS pairs were obtained in the same imaging session with CTCS obtained 5 minutes prior to any contrast injection. The exclusion criteria for the patients were: (1) age<20 years, (2) coronary artery bypass grafting, (3) acute or old myocardial infarction, (4) complete left bundle branch block, and (5) inadequate datasets such as poor image quality of CCTA. Of the 83 patients, 14 had a MACE outcome, which was defined as cardiovascular death, acute myocardium infraction or revascularization. CTCS Agatston score was calculated using the conventional detection criteria (3 connected voxels >130 HU) and summed to give a total score.

**Image processing and feature extraction from PCAT**

To analyze PCAT in paired CCTA and non-contrast CTCS images, several image processing steps were performed. Using the CCTA images with ready visualization of coronary arteries, experts semi-automatically segmented and identified centerlines in the three main coronary arteries (LAD, LCX, and RCA), using commercial software (Intuition Client version 4.4.13.P7, TeraRecon, Inc). For CCTA PCAT evaluation, regions determined to be of interest were obtained from a manually determined start point to help automatically identified end point a standard distance along the centerline (e.g., 40-mm). Since coronaries are poorly visible in non-contrast CTCS images, we registered CCTA (floating image volume) to CTCS (reference image volume) to identify the coronaries. We used the Deeds non-rigid registration method (19), which uses a minimum spanning tree approach to find a global optimum. This method has been found to be robust to "sliding" organs, a desirable attribute for heart registration. Prior to registration, we sampled CCTA images to be the same size of CTCS images and windowed the HU range of both CCTA and CTCS images (-300 HU, 300 HU) to give better contrast for the coronary artery regions. After registration, masks for the main coronary arteries (LAD, LCX and RCA) were transformed to corresponding locations in CTCS images.

We developed a novel axial-disk method for analysis of PCAT rather than the more conventional curved multiplanar reformatting (MPR) to avoid the potential confound from interpolating thick (3-mm) image slices in CTCS. We created axial-disk-masks centered on the coronary artery center with 2-times the axial diameter of the first 40-mm segmented lumen, giving binary axial-disk-masks. For RCA, we excluded the first 10-mm segment of the artery to avoid interference from the aortic wall. To ensure a reasonable PCAT size, the maximum diameter of the disk is limited to 8-mm. Following noise reduction with a 3x3 median filter, we apply the fat HU window [-190, -30] to identify PCAT (Figure 1).

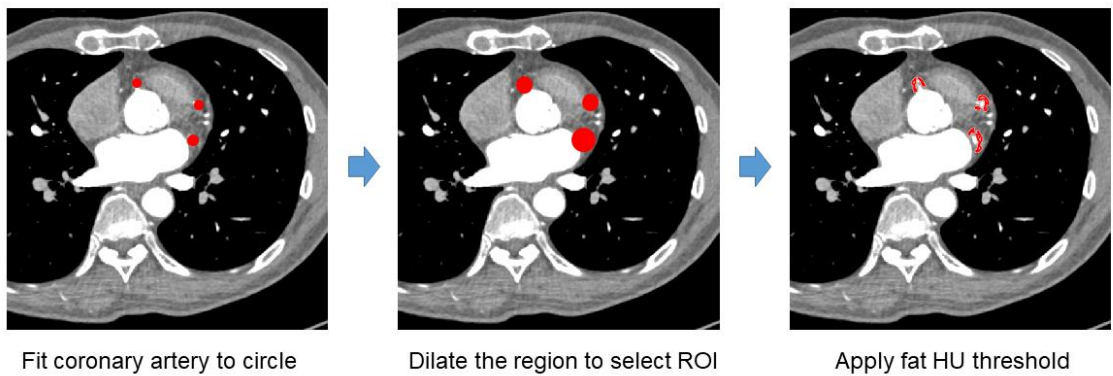

Figure 1. "Axial-disk" method for pericoronary adipose tissue (PCAT) assessment. We first fitted coronary artery segmentation in axial slice to a circle (red circles in left figure), giving out its center and radius, then dilated the region to 2-times the diameter (red circles in the middle, not exceeding 8-mm) and applied fat HU threshold to give out the final PCAT region (red contours in right figure).

PCAT features were extracted from voxels within binary masks in CCTA and CTCS axial-disk regions. From PCAT, we extracted 22 hand-crafted and 536 radiomics library (Pyradiomics (20)) features. For hand-crafted features, we focused on the HU histogram feature of PCAT HU values (e.g., small histogram, skewness, and kurtosis of the HU value histogram) and the PCAT axial area features (e.g., min and max area of PCAT in the axial disk). For radiomic features, we extracted: 1) shape features like surface volume ratio and major axis length; 2) texture features like Gray-Level Co-Occurrence Matrix (GLCM), Gray Level Dependence Matrix (GLDM), Gray Level Run Length Matrix (GLRLM), and Neighborhood Grey-Tone Difference Matrix (NGTDM), among many that have been used in CCTA analysis (21–23). Texture features were calculated using PCAT voxels with 16 bins of discretization. Radiomic were extracted at both original PCAT images and after three-dimensional wavelet transformation. Wavelet transformation decomposes the data into high and low-frequency components, enabling capturing discontinuities, ruptures and singularities and coarse structure of the data.

As registration quality was important to the pipeline, we evaluated registration quality visually and quantitatively. To quantify registration results, we manually segmented coronary arteries in CTCS images and compared results to the registered arteries from CCTA. Distance errors were assessed by the average distance between the centers of the axial disks in registered coronary artery masks originating from CCTA and the manually segmented ones from CTCS.

**Machine learning and statistical analyses**

For both CCTA and CTCS images, we determined the relative importance of various PCAT features and made MACE predictions. We compared the same features in paired CTCS (contrast free) and CCTA (with iodine contrast) using Spearman's rho correlation. We used univariate logistic regression to generate p-values for each hand-crafted and radiomic feature in Manhattan plots [-log10(P-values)]. Bonferroni correction to reduce false positives was applied to features by dividing the significance level of $\alpha = 0.05$ by the number of components that described 99.5% of the radiomic variation (24). We used multivariate logistic regression and five-fold cross-validation to build MACE predictive models and compared the area under the receiver operating characteristic curves (AUC) to assess performance on testing data. In addition, we created a combined model using bagging strategy which takes the maximum probability of all three coronary arteries.

**Results**

Registration accuracy was sufficient to enable the identification of PCAT regions in CTCS images (Figure 2). Across patients, the average distance between the centers of the axial disks generated in CCTA and registered to CTCS was only 1.36 ± 0.78 mm from those manually marked in CTCS images. This is comparable to the size of a voxel (0.4-mm x 0.4-mm x 3-mm voxels). As PCAT is segmented using the fat-window thresholds within relatively large axial disks (<8-mm diameter), this uncertainty is deemed acceptable.

Streak artifacts characteristic of beam hardening and/or motion were commonly observed in the CCTA PCAT regions (Figure 3). Compared to these artifact-containing CCTA images, CTCS images were homogeneous in the PCAT region. Images devoid of iodine have greatly reduced beam hardening due to the absence of iodine in the ventricle and reduced motion artifacts due to the absence of high-contrast moving objects.

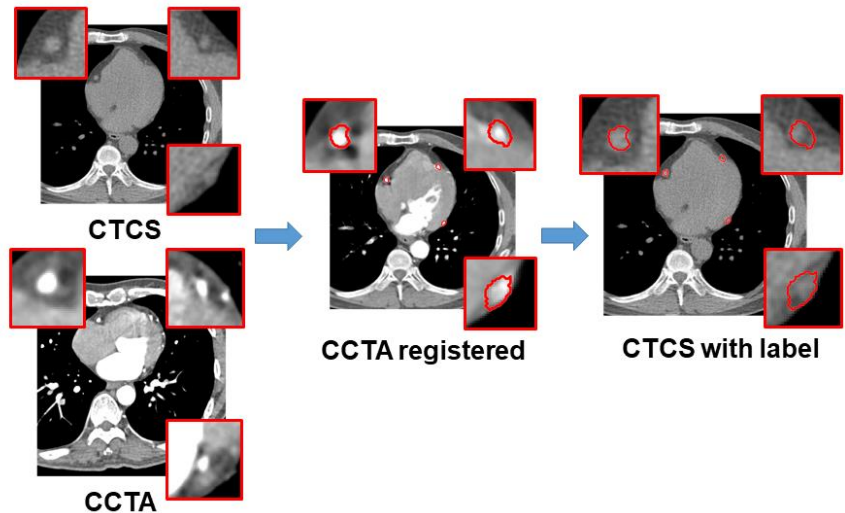

Figure 2. Segment coronary arteries in CT calcium score (CTCS) using registration results in coronary CT angiography (CCTA). After registering CCTA images to CTCS images, CCTA semi-automatic coronary artery segmentation was deformed (shown as red contours in the middle) and copied to CTCS (shown as red contours on the right). We could see registered segmentations aligned well with images evidence in CTCS, indicating feasibility to assess pericoronary adipose tissue (PCAT) in non-contrast CTCS images.

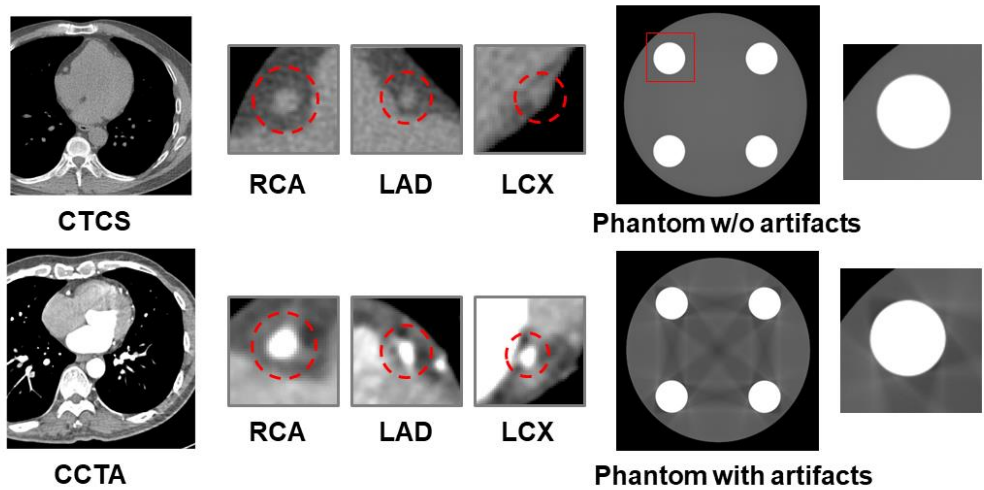

Figure 3. Artifacts in CCTA PCAT greatly affect PCAT features. The upper ones are CTCS image and its corresponding PCAT candidate regions while the lower ones for CCTA. In zoomed images, red dashed circles are the candidate regions before applying fat threshold. We could observe streaking artifacts in CCTA PCAT candidate regions, but they were non-existent in homogeneous CTCS images without iodine in the coronaries and ventricular cavity to give beam hardening and accentuate motion artifacts. We plotted the original phantom mono-energetic (upper) and poly-energetic reconstruction which shows beam hardening artifacts (lower). The zoomed phantom images are from upper left insert (red rectangle, 110 HU), the one with beam hardening artifact showed similarity to CCTA zoomed PCAT regions. Abbreviations as in Figure 2.

In Figure 4, for both CTCS and CCTA, we collected territory-specific mean HU and volumes for MACE and no-MACE patient groups. The analysis of the mean HU and volumes in the MACE group revealed slightly higher values in CTCS/CCTA scans, with respective mean differences of 3.19/1.15 HU and 0.46/0.2 $cm^3$. It is noteworthy that these differences were positive and comparable in both CTCS and CCTA images.

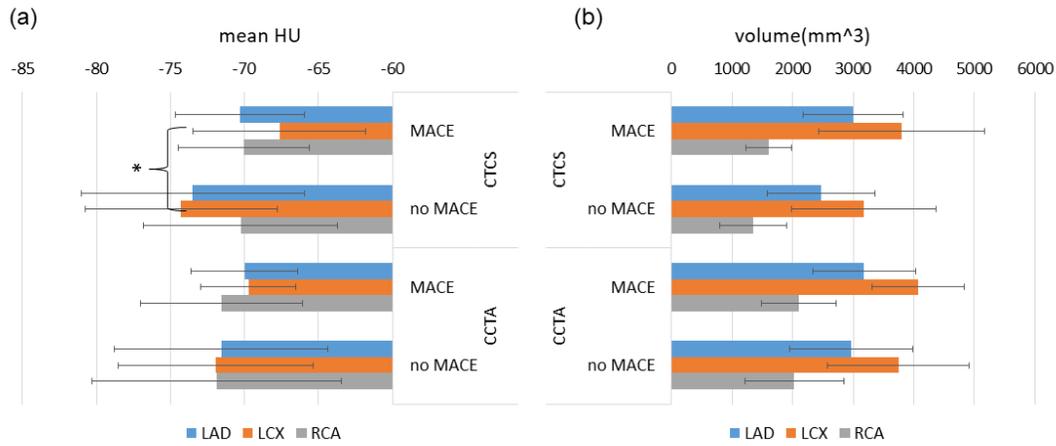

Figure 4. Comparison in PCAT mean HU and volume in MACE and no-MACE group. We observed slightly higher mean HU value (a) in MACE group in both CTCS and CCTA for three territories. The star indicated CTCS LCX mean HU value showed significant difference between MACE and no MACE group using student t-test. Similar trend was observed as MACE group had larger PCAT volumes in (b). Abbreviations as in Figure 2.

We extracted and analyzed hand-crafted features from PCAT in CTCS and CCTA images. Figure 5 depicts a Manhattan plot showing the strength of association between individual PCAT hand-crafted features and MACE, as obtained using univariate logistic regression. In general, CTCS PCAT features outperformed those from CCTA, with more features deemed discriminative based at all conventional and Bonferroni-adjusted significance levels. In Figure 6, we examine correlations between CTCS and CCTA features using a heatmap of Spearman's rho correlation. We observed good correlation in "axial area" features (mean rho=0.61) between CTCS and CCTA, indicating that CTCS could achieve similar PCAT morphology compared to CCTA. Notably, there was also correlation in mean HU and the probability of being in the most elevated HU histogram bin [-50 HU, -30 HU], the bin presumably most related to PCAT inflammation. By employing a multivariable logistic model and incorporating hand-crafted features, we assessed the predictive capability for determining MACE (Figure 7). Distinctly, the CTCS AUC was better than that for CCTA (0.83 versus 0.79), although we cannot reliably reject the null hypothesis of no difference (p=0.57). Nevertheless, this preliminary result suggests that CTCS may be as good or better than CCTA for this analysis. In addition, although the CTCS PCAT fat features are quite different than coronary calcifications giving direct assessment of atherosclerosis, MACE prediction from CTCS PCAT was comparable to that for Agatston score, suggesting good promise.

In addition, we similarly extracted and analyzed radiomics library features from PCAT in CTCS and CCTA images (Figure 8-10). In Figure 8, once again, the CTCS PCAT features tended to outperform those from CCTA, with more features deemed discriminative based at all conventional and Bonferroni-adjusted significance levels. In Figure 9, we further investigated the best 20 CCTA PCAT radiomics features and examined their correlation with measurements from CTCS images. We found that "wavelet intensity radiomics" showed good correlation between CTCS and CCTA, while "wavelet texture radiomics" showed poorer correlation. In Figure 10, we combined the same PCAT features to predict MACE. Again, combining the vessels into a multi-instance combination was helpful. In this figure, CTCS, CCTA, and Agatston had AUCs of 0.83, 0.77, and 0.73, respectively. Although none were statistically different due to the small sample size, this preliminary result certainly suggests that there is value in a PCAT analysis of CTCS images. Interestingly, when we combined PCAT hand crafted and radiomics library features from CTCS, the AUC was slightly degraded (0.83 to 0.77) but this anomaly was insignificant.

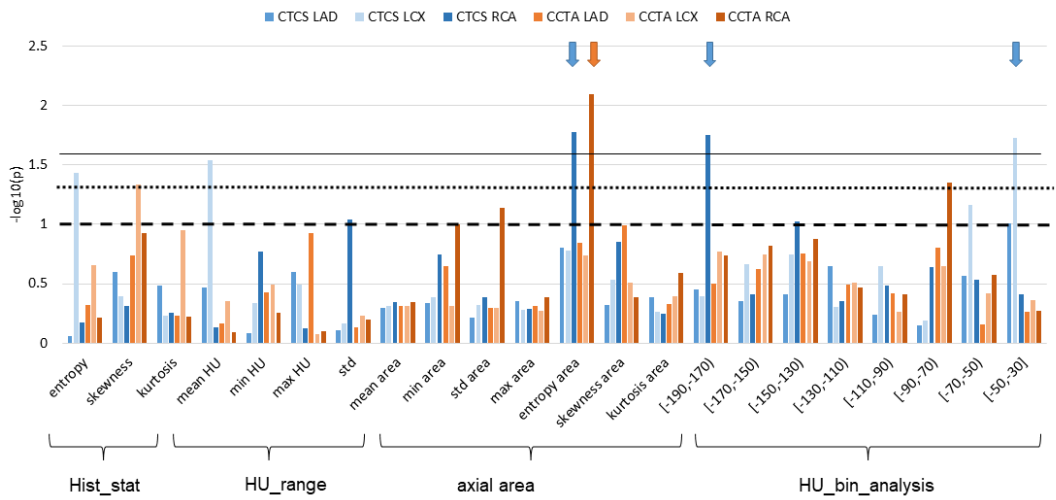

Figure 5. MACE informative hand-crafted PCAT features from CCTA and CTCS. In 3 coronaries, we extracted 22 features (3 histogram statistics, 4 histogram, 7 PCAT areas in axial slices, and 8 probability in ranges of HU values). Univariate logistic regression on MACE and no MACE groups gave p values. The dashed, dotted, and solid lines represent p=0.1, p=0.05, and the Bonferroni-adjusted significance level for p=0.05 (giving p=0.025), respectively. By raising the threshold to p=0.1, we can ensure the identification of discriminative features by limiting the number of false negatives. CTCS (blue)/CCTA (orange) gave 9/5, 5/3, and 3/1, for p equal to 0.1, 0.05, and 0.025, respectively. In all cases, the numbers of CTCS features deemed discriminative exceeded those of CCTA. Abbreviations as in Figure 2.

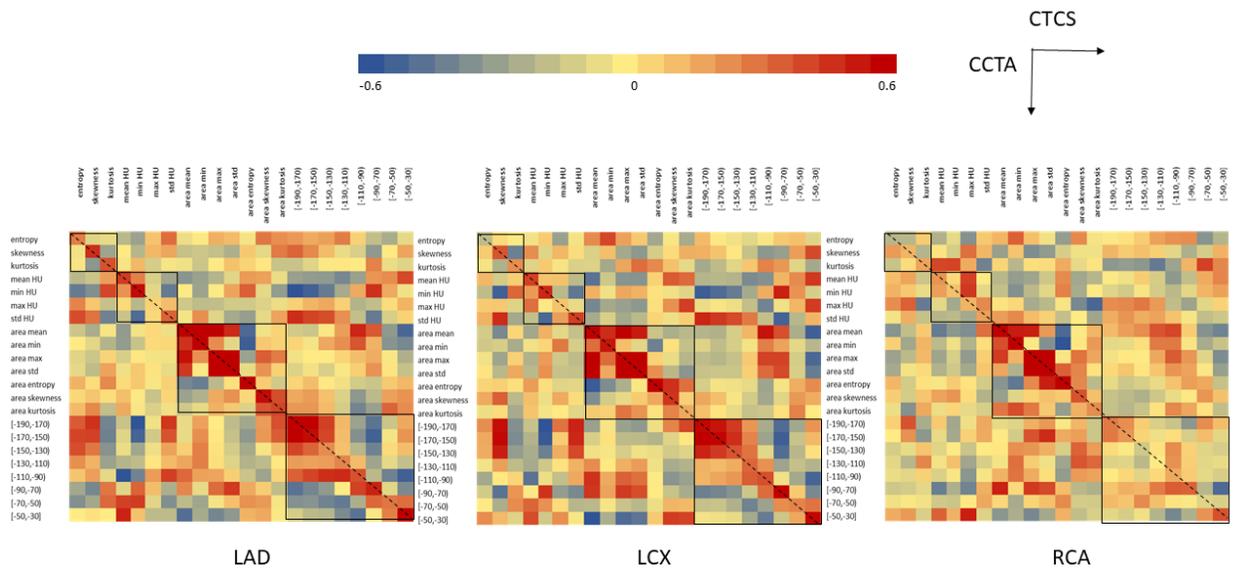

Figure 6. Correlation of PCAT features from CTCS and CCTA images. The orders for CTCS and CCTA PCAT features are the same in Spearman's rho heatmaps of three coronary arteries , the diagonal dashed line shows the correlation of the same feature across image modalities and boxes stand for each feature category derived from manual clustering. The best correlations were at "axial area" category, showing CTCS could catch similar PCAT morphology comparing to CCTA. Abbreviations as in Figure 2.

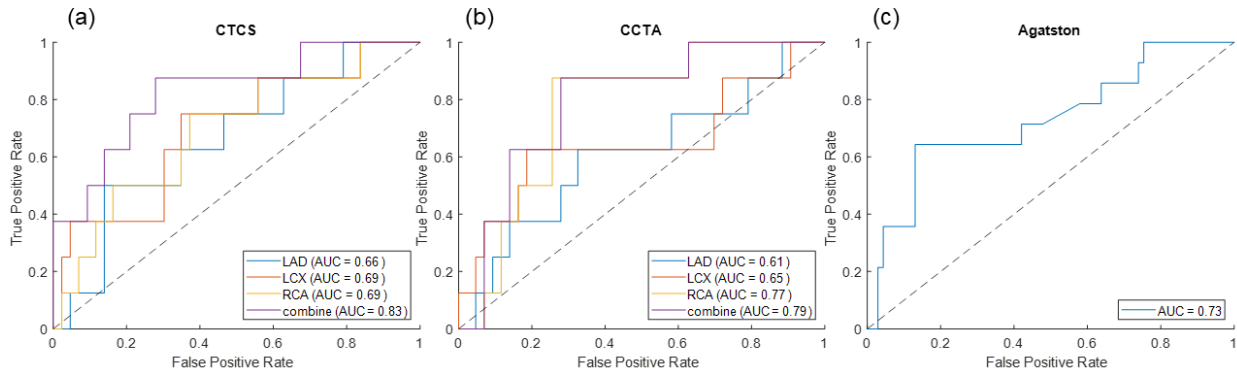

Figure 7. Prediction of MACE from PCAT handcrafted features extracted from CTCS and CCTA images. For CTCS, we show MACE prediction for features from individual arteries and for a combined multi-instance method where we take the maximum probability of MACE from each artery for a patient (a). In (b), we present a similar figure for CCTA. In (c), we show MACE prediction for the Agatston score. The multi-instance combined model gave the best result. In terms of best AUCs in each panel, CTCS, CCTA, and Agatston gave 0.83, 0.79, and 0.73, with no statistical difference between them. Abbreviations as in Figure 2.

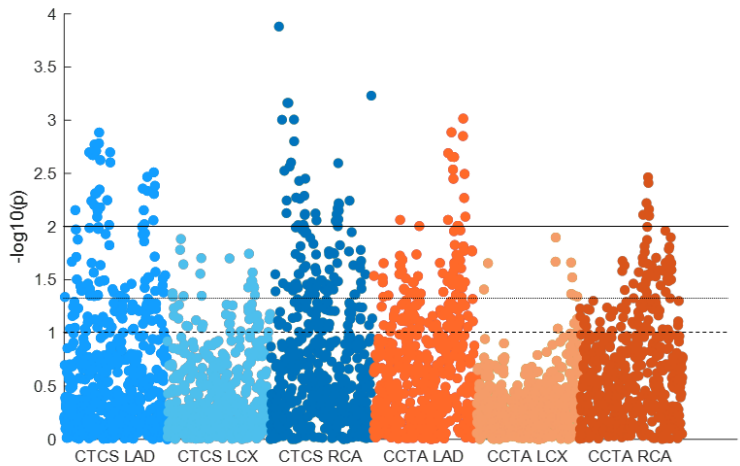

Figure 8. CTCS PCAT radiomics performed better than CCTA. We extracted 536 radiomic features on each coronary artery: 14 shape features, 40 texture features, 18 intensity features, 320 wavelet texture features, 144 wavelet intensity features. Univariate logistic regression on MACE and no-MACE groups gave p values. The dashed, dotted, and solid lines represent p=0.1, p=0.05, and the Bonferroni-adjusted significance level for p=0.05 (giving p=0.01), respectively. By raising the threshold to p=0.1, we can ensure the identification of discriminative features by limiting the number of false negatives. CTCS (blue)/CCTA (orange) gave 341/256, 203/137, and 63/22, for p equal to 0.1, 0.05, and 0.01, respectively. In all cases, the numbers of CTCS features deemed discriminative exceeded those of CCTA, indicating that CTCS PCAT features can achieve comparable performance compared to CCTA ones. Abbreviations as in Figure 2.

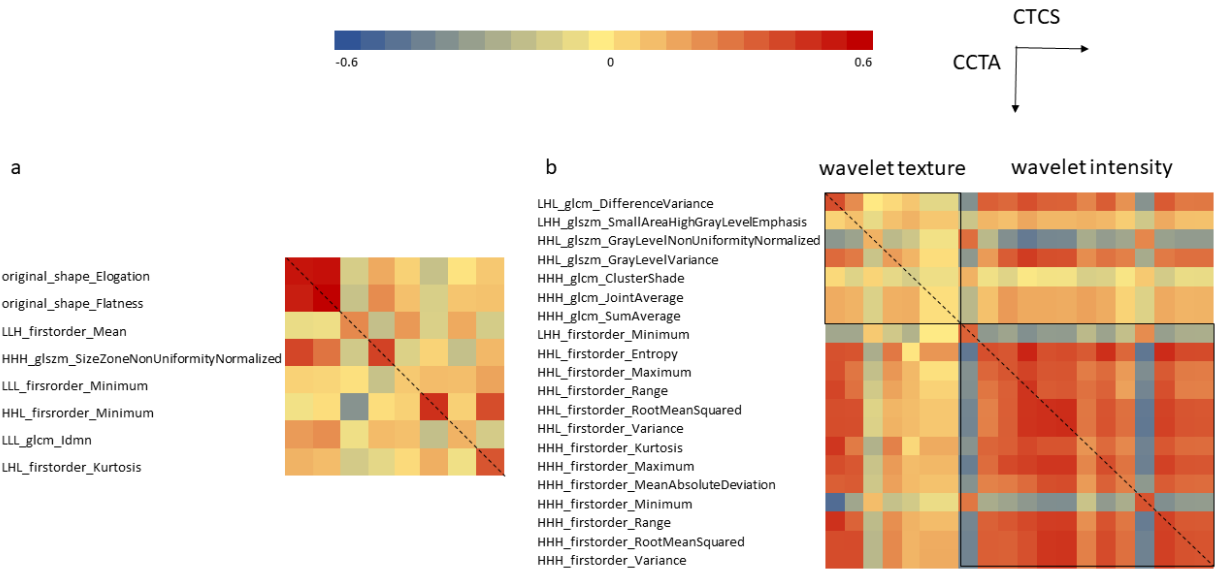

Figure 9. PCAT radiomics correlation in CTCS and CCTA. We take LAD for example, (a) shows left in Spearman's rho heatmap of reported good CCTA PCAT features in CTCS and CCTA, (b) is the best 20 CCTA PCAT radiomics in our study and its correlation to CTCS. The diagonal dashed line shows the correlation of the same feature across image modalities and boxes stand for each feature category. We observed shape related radiomics in (a) were highly correlated, and wavelet intensity radiomics in (b) were well correlated across image modalities, showing CTCS could catch similar PCAT morphology comparing to CCTA. However, wavelet texture radiomics of CCTA PCAT were not similar to CTCS ones, which can be visually seen in Figure 3. Abbreviations as in Figure 2.

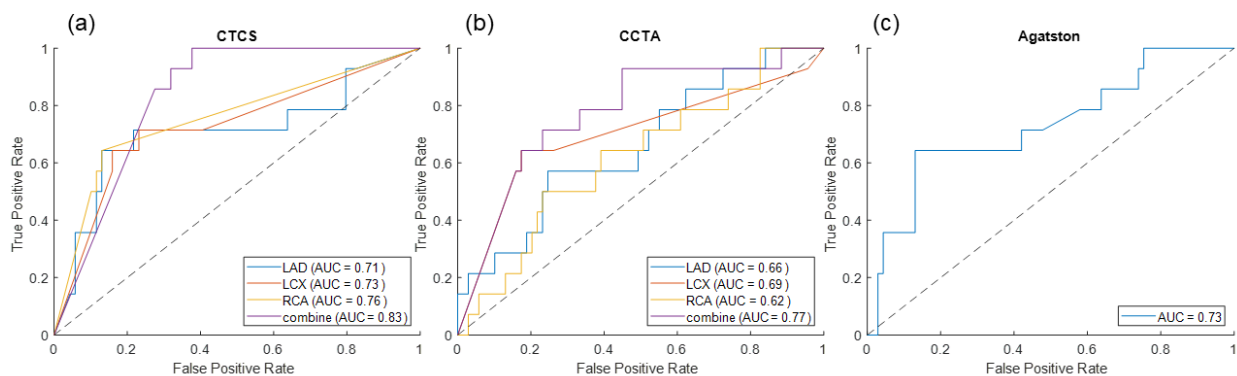

Figure 10. Prediction of MACE from PCAT radiomic features extracted from CTCS and CCTA images. For CTCS, we show MACE prediction for a combined multi-instance method from each artery performed best(a). In (b), we present a similar figure for CCTA. In (c), we show MACE prediction for the Agatston score. In terms of best AUCs in each panel, CTCS, CCTA, and Agatston gave similar performance with no statistical difference between them. Abbreviations as in Figure 2.

## Discussion

In this study, we investigated the feasibility of utilizing non-contrast CTCS images for PCAT assessment. Some of the main finding follow. Registration with CCTA images enabled, for the first time, the identification of PCAT regions from CTCS images from all three major coronary arteries. Univariate analysis provided a number of promising features from CTCS images for MACE prediction that actually exceeded the number in paired CCTA images. Some hand-crafted features and radiomics library features showed correlation between CTCS and CCTA. For prediction of MACE, CTCS PCAT features slightly outperformed features from CCTA, albeit not in a signifi-cant way with our limited paired dataset. CTCS PCAT analysis exhibits comparable or near-comparable performance to that of CCTA. Taken together, results suggest promise in PCAT analysis from CTCS images.

Using our processing pipeline, we were able to automatically localize PCAT in three major coronary arteries in CTCS images. Existing studies on non-contrast images have extracted PCAT assessments using manual segmentation of the RCA artery (17,18). The registration error was small (~1.4-mm), comparable to voxel sizes (Figure 2). Registration quality was deemed sufficient to segment PCAT. Because axial disks tended to be about 6-mm in diameter, small displacements of the axial disk center would have a small effect on results. In fact, when we evaluated PCAT RCA volumes from CTCS images using our pipeline, we found good agreement (Dice score 0.83±0.11, N=5) with that from manual identification of the vessel center. As argued previously, a curved MPR PCAT analysis common for 0.75-mm thick CCTA images would be inappropriate for 3-mm thick CTCS images, due to extreme oversampling of voxels in regions of high vessel curvature. We consider the axial-disk method the most appropriate alternative.

Regarding PCAT, there are notable differences between CTCS and CCTA, which may favor using CTCS. In a previous report using CT cardiac perfusion imaging (15), we identified the presence of iodine perfusion in PCAT and demonstrated changes in PCAT HUs, apparent volumes, and radiomics due to the presence of the iodine bolus in PCAT. This could represent a major confounding factor in CCTA images across individuals due to individual hemodynamic and CCTA acquisition time. In addition, beam hardening and motion artifacts prevalent in CCTA images could further result in marked image artifacts that may degrade the quality of CCTA images for PCAT evaluation (Figure 3). The non-contrast CTCS images have much more homogenous PCAT regions. These issues pose major advantages of CTCS over CCTA.

Regardless of their differences, the trends observed in PCAT assessment were consistent between CTCS and CCTA. In both CTCS and CCTA images, we observed higher mean HU value and larger volume of PCAT are likely to associate with MACE (Figure 4), similar to previous reports (8,9). In addition, interesting PCAT intensity features like mean HU and PCAT HU range [-50, -30] showed good correlation between CTCS and CCTA. Similarly, we also found good correlation between CCTA and CTCS for MACE informative shape elongation and mean value after three-dimensional wavelet transformation. We believe CTCS could serve as an alternative method to study PCAT since assessments are in good correlation.

For both CCTA and CTCS PCAT radiomic studies, we found features extracted after wavelet transformation are more informative than ones from original image (Figure 9b). The literature supports the usefulness of wavelet features in CCTA, indicating possible correlation to fibrosis and vascularity, reflecting permanent changes in adipose tissue induced by chronic coronary inflammation (22,25,26). Interestingly, wavelet features in non-contrasted CTCS were also predictive, perhaps suggesting this to be a finding worthy of more investigation.

The principal limitation of this study is the limited sample size. The size was limited due to the difficulty in registering data and identifying PCAT regions in CTCS images. Nevertheless, this preliminary study is most promising suggesting that it will be a worthy endeavor to develop means for automatic analysis of PCAT in CTCS images.

In conclusion, our study demonstrates that CTCS images can be used to analyze PCAT and that PCAT assessments in CTCS might be as valuable for predicting cardiovascular risk prediction as those from CCTA. Importantly, CTCS images do not suffer from the confounds found in CCTA images due to the presence of iodine. Given the low cost and prevalence of CTCS images, additional studies of PCAT in CTCS images are warranted.

**Clinical perspective**

Pericoronary adipose tissue attenuation, as evaluated in CT angiography, has been linked to fat inflammation, coronary artery disease, and cardiovascular risk. Our study reveals the feasibility and advantage of assessing pericoronary adipose tissue (PCAT) from non-contrast CT calcium score images, with extracted features that can predict MACE. CT calcium score images are not hindered by the presence of iodine and artifacts found in CT angiography images. Results suggest that screening, low-cost CT calcium score images could be an alternative for the assessment of PCAT. Future studies are required to validate the generalizability of our findings, which hold promise for enhancing protocols in assessing cardiovascular risk.

**Acknowledgments**

This research was supported by National Heart, Lung, and Blood Institute through grants R01HL167199, R01HL165218, R01 HL143484, and R44HL156811. This research was conducted in space renovated using funds from an NIH construction grant (C06 RR12463) awarded to Case Western Reserve University. The content of this report is solely the responsibility of the authors and does not necessarily represent the official views of the National Institutes of Health. The grants were obtained via collaboration between Case Western Reserve University, University Hospitals Cleveland Medical Center, and Mackay memorial hospital. This work made use of the High-Performance Computing Resource in the Core Facility for Advanced Research Computing at Case Western Reserve University. The veracity guarantor, Hao Wu, affirms to the best of his knowledge that all aspects of this paper are accurate.